\documentclass{article} 
\usepackage{iclr2018_conference,times}
\usepackage{hyperref}
\usepackage{breakurl}
\usepackage{mathtools}
\usepackage[dvips]{graphicx}
\usepackage{subfigure}
\usepackage{dcolumn}
\newcolumntype{d}[1]{D{.}{.}{#1}}
\newcommand*{\around}{\mathord{\sim}}
\title{Improving Deep Learning by Inverse Square Root Linear Units (ISRLUs)}
\author{
Brad Carlile, Guy Delamarter, Paul Kinney, Akiko Marti, Brian Whitney \\
AI Perf Engineering \\
\texttt{info@aiperf.com}
}

\iclrfinalcopy 

\begin{document}

\maketitle\lhead{Under review as a conference paper 2018}

\begin{abstract}
  We introduce the ``inverse square root linear unit'' (ISRLU) to
  speed up learning in deep neural networks. ISRLU has better
  performance than ELU but has many of the same benefits. ISRLU and
  ELU have similar curves and characteristics. Both have negative
  values, allowing them to push mean unit activation closer to zero,
  and bring the normal gradient closer to the unit natural gradient,
  ensuring a noise-robust deactivation state, lessening the over
  fitting risk.  The significant performance advantage of ISRLU on
  traditional CPUs also carry over to more efficient HW
  implementations on HW/SW codesign for CNNs/RNNs.  In experiments
  with TensorFlow, ISRLU leads to faster learning and better
  generalization than ReLU on CNNs.  This work also suggests a
  computationally efficient variant called the ``inverse square root
  unit'' (ISRU) which can be used for RNNs. Many RNNs use either long
  short-term memory (LSTM) and gated recurrent units (GRU) which are
  implemented with $\tanh$ and sigmoid activation functions. ISRU has
  less computational complexity but still has a similar curve to
  $\tanh$ and sigmoid.
\end{abstract}

\section{Introduction}

Two popular activation functions for neural networks are the rectified
linear unit (ReLU) \citep{GlorotEtAl2011} and the exponential linear
unit (ELU) \citep{ClevertEtAl2015}. The ReLU activation function is the
identity for positive arguments and zero otherwise. The ELU activation
function is the identity for positive arguments and has an exponential
asymptotic approach to -1 for negative values.

From previous analysis of the Fisher optimal learning, i.e., the
natural gradient \citep{Amari1998,ClevertEtAl2015}, we can reduce the
undesired bias shift effect without the natural gradient, either by
centering the activation of incoming units at zero or by using
activation functions with negative values. We introduce the inverse
square root linear unit (ISRLU), an activation function like ELU, that
has smoothly saturating negative values for negative arguments, and
the identity for positive arguments. In addition this activation
function can be more efficiently implemented than ELU in a variety of
software or purpose-built hardware.

\section{Inverse Square Root Linear Unit (ISRLU)}

The {\em inverse square root linear unit} (ISRLU) with $\alpha$ is

\begin{align}
f(x) = \begin{cases}
x & \text{if $x \ge 0$} \\
x \left ( \frac{1}{\sqrt{1+\alpha x^2}} \right ) & \text{if $x<0$}
\end{cases} , \qquad
f'(x) = \begin{cases}
1 & \text{if $x \ge 0$} \\
\left ( \frac{1}{\sqrt{1+\alpha x^2}} \right )^3 & \text{if $x<0$}
\end{cases}
\end{align}

\begin{figure}[ht]
\begin{center}
\includegraphics[width=0.9\linewidth]{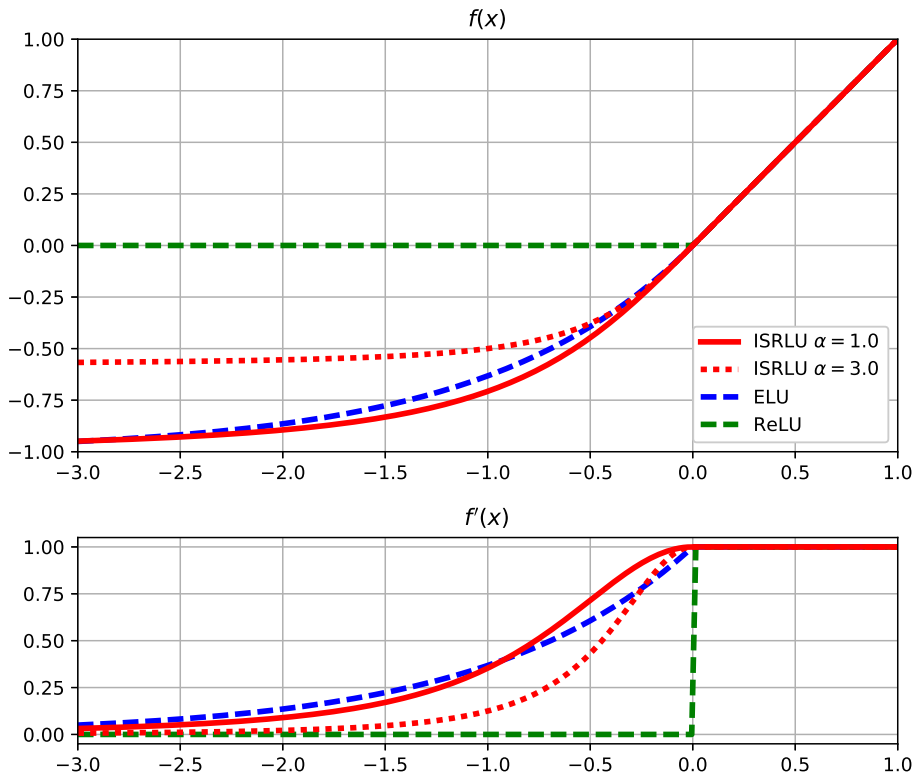}
\end{center}
\caption{The inverse square root linear unit (ISRLU),
  ISRLU ($\alpha=1$;$\alpha=3$), ELU ($\alpha=1$), and ReLU; and their first
  derivatives.}
\label{fig:activation-plots}
\end{figure}

The ISRLU hyperparameter $\alpha$ controls the value to which an ISRLU
saturates for negative inputs (see Fig.~\ref{fig:activation-plots}).
ISRLUs and ELUs have very similar curves so at a high level one would
expect to see the same general characteristics in most cases. ISRLUs
have smooth and continuous first and second derivatives. ELUs are only
continuous in the first derivative (see
Fig.~\ref{fig:activation-plots}). In contrast, ReLU is
non-differentiable at zero.  Since ISRLUs and ELUs share most of the
same characteristics we use the same weight initialization guidelines
as are used for ELUs \citep{ClevertEtAl2015}).

The primary advantage of ISRLU is in its reduced computational
complexity compared to ELU. Inverse square roots are faster to
calculate than exponentials. When calculating ISRLU for negative
inputs, first one calculates $1/\sqrt{1 + \alpha x^2}$. Multiplying
this function by $x$ provides the value for the forward calculation.
Multiplying this function by itself twice (i.e. cubing) provides the
value for back-propagation.

With $\alpha=1$, ISRLU saturation approaches -1. With $\alpha=3$, the
negative saturation is reduced, so a smaller portion of the
back-propagated error signal will pass to the next layer. This allows
the network to output sparse activations while preserving its ability
to reactivate dead neurons. Future work will establish what deeper
saturation ($\alpha<1$) is appropriate when applying ISRLU to
self-normalizing neural networks~\citep{KlambauerEtAl2017}. Note that
under variations of the $\alpha$ parameter, ISRLU is a family of
continuous curves which preserve their smooth derivatives, and
transition between the functional bounds of ReLU
($\alpha\rightarrow\infty$) and linear ($\alpha\rightarrow 0$).

In the same manner as parametric ReLUs (PReLUs) only one additional
hyperparameter is required and methods can be used to directly learn
its value during back-propagation \citep{HeEtAl2015}. Similarly, ISRLU's
$\alpha$ can be learned during the training phase along with the weights and
biases. Indeed for PReLUs, \cite{HeEtAl2015} have empirically shown
that learning the slope parameter ``a'' gives better performance than
manually setting it to a pre-defined value.

\section{Activation Function Performance}

\cite{ShahEtAl2016} showed that ELU was faster than the combination of
ReLU and Batch Normalization for deep neural network (DNN) ResNet
architectures. On CIFAR-10 and CIFAR-100 they showed that ELU not only
speeds up learning but also improves the accuracy as the depth of the
convolutional neural network (CNN) increases.

More than learning rate needs to be considered when evaluating the
overall performance of CNNs. The amount of time and computational
resources required to perform both the convolutions and activation
functions combined should be considered.

The trend in CNNs is that less time is being spent calculating
convolutions. There are three factors that we are seeing. First is
that small convolution filters such as 5x5 or 3x3 filters are the
basis of many architectures. Second, architectures as Inception-v3 and
Inception-v4 now decompose 2d filters such as a 3x3 into a 3x1 filter
and a 1x3 filter \citep{SzegedyEtAl2016}. Third, more efficient
calculations of convolution that rely on techniques such as Winograd's
minimal filtering algorithm \citep{LavinGray2016, Winograd1980} are
being used for 3x3 and smaller filters as are FFTs to reduce
calculation time in 5x5 or larger filters. All of these techniques
reduce the amount of calculations for each element in the convolution
output.  

Table~\ref{tbl:filter-complexity} shows ``cycles per output element''
for an Intel Xeon Platinum 8160 (Skylake).

\begin{table}[h]
\caption{Computational complexity of various filter sizes}
\label{tbl:filter-complexity}
\begin{center}
\begin{tabular}{ld{2.0}d{2.0}d{1.2}}
\multicolumn{1}{c}{\bf Convolution}  &
\multicolumn{1}{p{0.15\linewidth}}{\centering \bf FP \\ Multiplies}   &
\multicolumn{1}{p{0.15\linewidth}}{\centering \bf FP \\ Adds} &
\multicolumn{1}{p{0.25\linewidth}}
{\centering \bf Cycles per output \\ element (CPE)}
\\ \hline \\
5x5                       &   25       &  24  & \around 4.25 \\
3x3                       &    9       &   8  & \around 1.53 \\
3x1, 1x3 Inception-v3, -v4&    3       &   2  & \around 0.51 \\
\end{tabular}
\end{center}
\end{table}

Due to all of these reductions in convolution computational
complexity, activation function performance is now a greater part of
overall learning performance.

Another characteristic that is changing with the use of smaller
filters is the decrease in the compute intensity
\citep{Carlile1993a,Carlile1993b}, which raises the importance of
memory systems performance for CNNs. The compute intensity of an
algorithm is the ratio of the number of operations divided by number
of words accessed. For a given algorithm it is straightforward to
calculate the upper bound of the computation rate that can be
supported on a given memory bandwidth.

\subsection{Activation Function Implementation} 

The main advantage of ISRLU over ELU is that it is based on the
inverse square root, which has been faster to evaluate than the
exponential for many generations of systems. In the past, whenever it
has not been faster, optimization potentials for inverse square root
implementation improvement have been found. It is instructive to
understand the current CPU performance of the inverse square root
intrinsic performance compared to exponentials and $\tanh$.

Intel x86 CPUs with SIMD instructions have vector intrinsic functions
to accelerate performance. Intel publishes CPE (Clocks per Element)
for various vector functions on their ``Vector Mathematics (VM)
Performance and Accuracy Data'' website, see
Table~\ref{tbl:CPU-performance} \citep{IntelPerformance}.

\begin{table}[h]
\caption{CPU performance on vector inverse square root, Exp, Tanh (x86).}
\label{tbl:CPU-performance}
\begin{center}
\begin{tabular}{ld{1.2}d{1.2}d{1.2}}
\multicolumn{1}{p{0.2\linewidth}}
{\centering \bf Vector Function \\ Single Precision (EP) } &
\multicolumn{1}{p{0.2\linewidth}}
{\centering \bf Intel Xeon E5-2699 v3 \\ (Haswell AVX2) } &
\multicolumn{1}{p{0.2\linewidth}}
{\centering \bf Intel Xeon E5-2699 v4 \\ (Broadwell AVX2)} &
\multicolumn{1}{p{0.21\linewidth}}
{\centering \bf Intel Xeon Platinum 8180 \\ (Skylake AVX-512) }
\\ \hline \\
InvSqrt      & 0.66 & 0.64 & 0.24 \\
Exp          & 0.81 & 0.89 & 0.52 \\
Tanh         & 4.19 & 4.43 & 0.78 \\
\noalign{\medskip}
\hline
\noalign{\medskip}
Exp/InvSqrt  & 1.2\times & 1.4\times & 2.2\times \\
Tanh/InvSqrt & 6.3\times & 6.9\times & 3.3\times 
\end{tabular}
\end{center}
\end{table}

For example, on a 3x1 filter using ELU in the negative region,
approximately the same CPE is required to evaluate the convolution as
is required for the exponential (cf. Table~\ref{tbl:filter-complexity} and
Table~\ref{tbl:CPU-performance}). Improvements in activation function
performance will impact overall time spent in each learning step.

We measured the vector performance of AVX2 implementations for the
various activation functions. The dataset used was 50\% negative and
50\% positive.  Results are shown in Table~\ref{tbl:AVX2-performance}.

\begin{table}[hb]
\caption{Vector ISRLU, ISRU, ELU, and ReLU performance on AVX2 (Intel Core i7-7700
Processor [$3.60\;\textrm{GHz}$ ``Kaby Lake''] ).}
\label{tbl:AVX2-performance}
\begin{center}
\begin{tabular}{ld{1.3}d{1.2}d{1.2}}
\multicolumn{1}{p{0.3\linewidth}}
{\centering \bf Activation Function \\ Single Precision} &
\multicolumn{1}{p{0.2\linewidth}}
{\centering \bf nsec/ \\ element} &
\multicolumn{1}{p{0.2\linewidth}}
{\centering \bf ISRLU \\ Perf Advantage} &
\multicolumn{1}{p{0.2\linewidth}}
{\centering \bf ISRLU (approx.) \\ Perf Advantage}
\\ \hline \\
ReLU                   & 0.340  & 0.62\times & 0.99\times \\
ISRU (approx.)         & 0.334  & 0.61\times & 0.97\times \\
ISRLU (approx.)        & 0.344  & 0.62\times & 1.00\times \\
ISRLU                  & 0.551  & 1.00\times & 1.60\times \\
ELU                    & 1.447  & 2.63\times & 4.21\times 
\end{tabular}
\end{center}
\end{table}

These results show that ISRLU ($\alpha=1.0$) is $2.6\times$ faster than
ELU. The fast approximation of ISRLU is within $1\%$ of the evaluation
speed of ReLU while still retaining all of the desired learning curve
properties mentioned in this paper.  This fast approximation for ISRLU
on this processor has only $3\times10^{-4}$ maximum relative error
($\around 11.6$ accurate bits).  One Newton-Raphson iteration doubles
that to $\around 23.4$ accurate bits out of the 24 bits of mantissa, and
two iterations achieves full precision.  We plan to evaluate if the
fast approximation has similar learning rates of the full precision
ISRLU.

\subsection{A Practical Trick for Inverse Square Root Calculation}

It is instructive to look at a practical trick for the computation of
the inverse square root as it may serve as inspiration for those
implementing ISRLU in hardware. Software implementations on CPUs can
take advantage of floating-point formats for faster evaluation of the
inverse square root. John Carmack and Terje Mathisen are often
associated with implementing fast inverse square root in 2002
\citep{Lomont2003}. In 1986, one of the authors of this paper
originally invented this method, which was called \ificlrfinal ``The
Kinney Method,'' \else ``The K Method,'' \fi to implement vector square
root for the production FPS T Series Hypercube Supercomputer
\citep{Gustafson1986}.  William Kahan and K.C. Ng at Berkeley also
independently discovered this around 1986.

Carmack \& Mathisen only used one iteration of the Newton method after
their fast approximation. One iteration had an error of approximately
0.175\%, which was suitable for their graphics applications. Since
various piecewise functions have been used to approximate activation
functions for CNNs and RNNs, part of our future research will look
into if fast approximations to ISRLUs are suitable for DNNs.

Another avenue to look at for hardware implementations of the inverse square
root is table-lookup hardware. Our expectation is that an efficient
hardware approximation for the inverse square root should take about the
same execution time as a fused multiply and add (FMA).  

\section{Experiments Using ISRLUs}

We used TensorFlow \citep{AbadiEtAl2016} to train a CNN on the (Lecun)
MNIST dataset. We tested the MNIST gray images in 10 classes, 60k
train and 10k test.

The first CNN architecture (see Table~\ref{tbl:MNIST-loss-accuracy-1})
in our experiments used 28x28 input, a convolutional layer with 6x6
with 6 feature maps, a convolutional layer with 5x5 with 12 feature
maps, a convolutional layer with 4x4 with 24 feature maps, a fully
connected layer of 1176 hidden units, and a softmax output layer with
10 units. Only a full-precision ISRLU was used in these initial tests due
to time constraints.

Convolutional neural networks with ISRLUs ($\alpha=1.0$, $\alpha=3.0$), ELUs
($\alpha=1.0$), and ReLUs were trained on the MNIST digit
classification dataset while each hidden units activation was tracked.
Each network was trained for 17 epochs by using ADAM optimizer with
learning rate 0.003 exponentially decreasing to 0.0001 and
mini-batches of size 100. The weights have been initialized to
truncated normal with standard deviation 0.1. The training error of
ISRLU networks decreases much more rapidly than for the other
networks. We also calculated the final cross-entropy loss function for
each test.

\begin{table}[h]
\caption{Architecture~1 on MNIST with test accuracy and cross-entropy loss with different activation functions.}
\label{tbl:MNIST-loss-accuracy-1}
\begin{center}
\begin{tabular}{ld{1.2}d{2.2}d{1.3}}
\multicolumn{1}{p{0.25\linewidth}}{\centering \bf Activation \\ Function}  &
\multicolumn{1}{p{0.1\linewidth}}{\centering \bf DropOut \\ pkeep} &
\multicolumn{1}{p{0.2\linewidth}}{\centering \bf Max Test \\ Accuracy} &
\multicolumn{1}{p{0.25\linewidth}}{\centering \bf Cross-Entropy \\ Loss}
\\ \hline \\
ISRLU $\alpha=3.0$ & 0.25 & 99.30 & 2.308 \\
ELU                & 0.40 & 99.29 & 2.395 \\
ISRLU $\alpha=3.0$ & 0.40 & 99.27 & 2.530 \\
ReLU               & 0.40 & 99.22 & 2.644 \\
ISRLU $\alpha=1.0$ & 0.40 & 99.20 & 2.785 \\
ReLU               & 0.25 & 99.17 & 2.798 \\
ELU                & 0.25 & 99.09 & 2.892 \\
ISRLU $\alpha=1.0$ & 0.25 & 99.00 & 3.124 
\end{tabular}
\end{center}
\end{table}

The second CNN architecture (see
Table~\ref{tbl:MNIST-loss-accuracy-2}) in our experiments used 28x28
input, a convolutional layer with 3x3 with 64 feature maps, a
convolutional layer with 3x3 with 64 feature maps, 2x2 Maxpooling,
DropOut, a convolutional layer with 3x3 with 64 feature
maps, a convolutional layer with 3x3 with 64 feature maps, 2x2
Maxpooling, DropOut, a fully connected (FC) layer of 512 hidden
units, and a softmax output layer with 10 units. Full-precision
ISRLU was used.

Convolutional neural networks with ISRLUs ($\alpha=1.0$, $\alpha=3.0$) and ELUs
($\alpha=1.0$) were trained on the MNIST digit classification dataset
while each hidden units activation was tracked. The network was
trained for 20 epochs by using ADAM optimizer with learning rate 0.003
exponentially decreasing to 0.0001 and mini-batches of size 100. The
weights have been initialized to truncated normal with standard
deviation 0.1.

\begin{table}[h]
\caption{Architecture~2 on MNIST with test accuracy and cross-entropy loss with different activation functions.}
\label{tbl:MNIST-loss-accuracy-2}
\begin{center}
\begin{tabular}{lccc}
\multicolumn{1}{p{0.25\linewidth}}{\centering \bf Activation \\ Function}  &
\multicolumn{1}{p{0.1\linewidth}}{\centering \bf DropOut \\ pkeep} &
\multicolumn{1}{p{0.2\linewidth}}{\centering \bf Max Test \\ Accuracy} &
\multicolumn{1}{p{0.25\linewidth}}{\centering \bf Cross-Entropy \\ Loss}
\\ \hline \\
ISRLU $\alpha=1.0$ & 
\parbox[t]{0.1\linewidth}{0.7 conv \\ 0.4 FC} &
99.32 & 2.334 \\[15pt]
ISRLU $\alpha=3.0$ & 
\parbox[t]{0.1\linewidth}{0.7 conv \\ 0.4 FC} &
99.30 & 2.389 \\[15pt]
ELU & 
\parbox[t]{0.1\linewidth}{0.7 conv \\ 0.4 FC} &
99.29 & 2.225
\end{tabular}
\end{center}
\end{table}

We did not expect significant differences in accuracy in ISRLU and ELU
in this test of shallow networks due to the similar nature of the
curves. The cross-entropy loss was reasonable, at between 2 and 3.2 for
all activation functions. Future testing will be done on deeper
networks where we expect larger advantages that are similar to ELU
\citep{ClevertEtAl2015,ShahEtAl2016}.

\section{Inverse Square Root Unit (ISRU)}

The work with ISRLU in this paper suggests that the {\em inverse square
root unit} (ISRU) may be useful for a variety of neural networks. ISRUs
are defined as:

\begin{align}
f(x) =
x \left ( \frac{1}{\sqrt{1+\alpha x^2}} \right )
, \qquad
f'(x) =
\left ( \frac{1}{\sqrt{1+\alpha x^2}} \right )^3
\end{align}

In RNNs that use LSTM \citep{HochreiterSchmidhuber1997} and GRU
\citep{ChungEtAl2014}, the most common activation functions are
sigmoid and $\tanh$. We assert that ISRUs can be more efficient
calculation than $\tanh$ and be more efficient than sigmoid when properly
shifted and scaled.  As shown above in Table~\ref{tbl:CPU-performance},
the inverse square root is 3x to 6x faster than $\tanh$ (depending on x86
architecture). ISRUs will be an area of our future research.

\begin{figure}[ht]
\begin{center}
\includegraphics[width=0.9\linewidth]{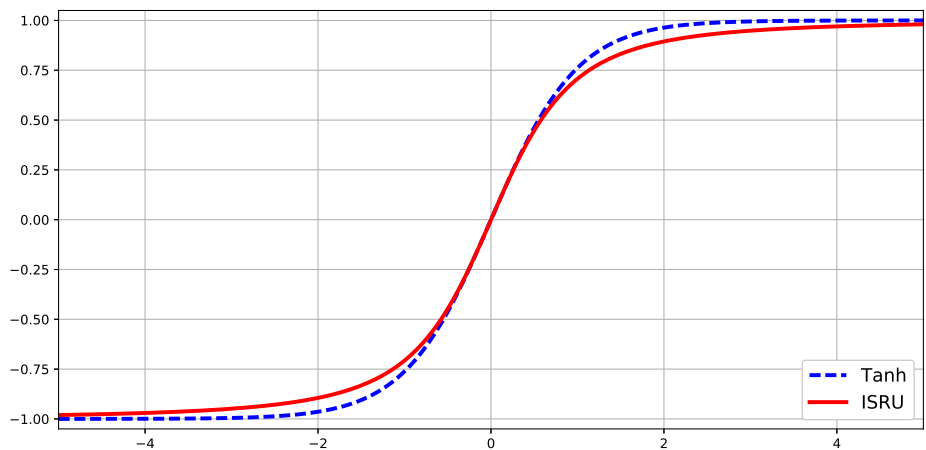}
\end{center}
\caption{The inverse square root unit (ISRU) and $\tanh$ functions.}
\label{fig:isru}
\end{figure}

\section{Conclusion}

Activation function performance is becoming more important overall in
convolutional neural networks (CNNs) because of the trending
reductions in the computational complexity of the convolutions used in
CNNs. We have introduced a new activation function, the inverse square
root linear unit (ISRLU) for faster and precise learning in deep
convolutional neural networks. ISRLUs have similar activation curves
to ELUs, including the negative values. This decreases the forward
propagated variation and brings the mean activations to zero. Mean
activations close to zero decreases the bias shift for units in the
next layer which speeds up learning by bringing the natural gradient
closer to the unit natural gradient. Future work may prove the
effectiveness of applying ISRLUs and the related ISRUs to other network
architectures, such as recurrent neural networks, and to other tasks,
such as object detection. ISRLUs have lower computational complexity
than ELUs. Even greater savings on computation can be realized by
implementing ISRLUs in custom hardware implementations. We expect
ISRLU activations to increase the training efficiency of convolutional
networks.

\bibliography{isrlu_iclr}
\bibliographystyle{iclr2018_conference}

\end{document}